\documentclass[sigconf]{acmart}

\usepackage{pdfpages}
\pdfoutput=1

\usepackage{booktabs} % For formal tables

\copyrightyear{2019}
\acmYear{2019}
\setcopyright{iw3c2w3}
\acmConference[WWW '19 Companion]{Companion Proceedings of the 2019 World Wide Web Conference}{May 13--17, 2019}{San Francisco, CA, USA}
\acmBooktitle{Companion Proceedings of the 2019 World Wide Web Conference (WWW '19 Companion), May 13--17, 2019, San Francisco, CA, USA}
\acmPrice{}
\acmDOI{10.1145/3308560.3316587}
\acmISBN{978-1-4503-6675-5/19/05}

\usepackage{balance}
\usepackage{booktabs} % For formal tables
\usepackage{color}
\usepackage{makecell}
\usepackage{pgfplots}
\usetikzlibrary{matrix}
\usepgfplotslibrary{groupplots}
\pgfplotsset{compat=newest}
\usepackage{tabularx}

\newcommand{\modelname}{\textit{MATAN} }

\begin{document}
\title{Link Prediction with Mutual Attention for Text-Attributed Networks}

\author{Robin Brochier}
\affiliation{%
  \institution{Universit\'e de Lyon, Lyon 2, ERIC EA3083}
  %\city{Lyon}
  %\state{France}
}
\email{robin.brochier@univ-lyon2.fr}

\author{Adrien Guille}
\affiliation{%
  \institution{Universit\'e de Lyon, Lyon 2, ERIC EA3083}
  %\institution{University of Lyon}
  %\city{Lyon}
  %\state{France}
}
\email{adrien.guille@univ-lyon2.fr}

\author{Julien Velcin}
\affiliation{%
  \institution{Universit\'e de Lyon, Lyon 2, ERIC EA3083 }
  %\institution{University of Lyon}
  %\city{Lyon}
  %\state{France}
}
\email{julien.velcin@univ-lyon2.fr}

\renewcommand{\shortauthors}{Brochier, Guille and Velcin}

\begin{abstract}
In this extended abstract, we present an algorithm that learns a similarity measure between documents from the network topology of a structured corpus. We leverage the Scaled Dot-Product Attention, a recently proposed attention mechanism, to design a mutual attention mechanism between pairs of documents. To train its parameters, we use the network links as supervision. 
We provide preliminary experiment results with a citation dataset on two prediction tasks, demonstrating the capacity of our model to learn a meaningful textual similarity. 
\end{abstract}




\maketitle

\section{Related Works}

In this section, we relate recent works in the fields of network embedding (NE) and attention mechanism for natural language processing (NLP).

\subsection{Attributed Network Embedding}

%NE consists in learning latent representations in a low-dimensional vector space of nodes in a network in order to facilitate tasks such as link prediction. 
\textit{DeepWalk} \cite{perozzi2014deepwalk} first proposed to derive the word embedding algorithm \textit{Word2vec} \cite{mikolov2013efficient} by generating paths of nodes, akin to sentences, with truncated random walks. \textit{DeepWalk} and other variants are generalized into a common matrix factorization framework in \textit{NetMF} \cite{qiu2018network}.
To extend \textit{DeepWalk} for text-attributed networks, \textit{TADW} \cite{yang2015network} expresses this latter as a matrix factorization problem and incorporates a matrix of textual features $T$, produced by latent semantic indexing (\textit{LSI}), into the factorization so that the vertex similarity matrix can be reconstructed as the product of three matrices $V^T$, $H$ and $T$. %We use \textit{TADW} as a baseline in our experiments and generate representations for unseen documents by using only its textual component $HT$. 

\subsection{Attention Mechanisms for NLP}

The \textit{Transformer} \cite{vaswani2017attention} is a novel neural architecture that outperforms state-of-the-art methods in neural machine translation (NMT) without the use of convolution nor recurrent units.
The Scaled Dot-Product Attention (SDPA) is the main constituting part of the \textit{Transformer} that actually performs attention over a set of words. It takes as input a query vector $q$ and a set of key vectors $K$ of dimensions $d_k$ and value vectors $V$ of dimensions $d_v$. One weight for a value is generated by a compatibility function with its corresponding key and the query. %The attention vector is generated as a weighted sum of the values with their associated attention weights. In practice, the attention vectors are generated in parallel for multiple queries $Q$, linked to the words of a document of length $L$, given the keys $K$ and the values $V$, linked to the words of another document. Formally, the attention vectors are produced with the following formula:
Formally, the attention vectors are generated in parallel for multiples queries $Q$, following the formula: $ \text{Attention}(Q,K,V)= \text{softmax}(\frac{QK^T}{\sqrt{d_k}})V$. 
The result is a set of $L$ attention vectors ($L$ being the number of queries) of dimension $d_v$. The matrices $Q$, $K$ and $V$ are produced by projection of initial words representations $W$ with three matrices $P^Q$, $P^K$ and $P^V$ whose parameters are meant to be learned.

Several works \cite{devlin2018bert, radford2018improving} adapted the \textit{Transformer} architecture beyond the task of NMT. Their main idea is to train the \textit{Transformer} in an unsupervised fashion over large corpora of texts and further refine its parameters on a wide variety of supervised tasks. Motivated by these recent works, we present a model, \modelname (\textbf{M}utual \textbf{A}ttention for \textbf{Te}xt-\textbf{A}ttributed \textbf{N}etworks), that derives from the the SDPA to address the task of link prediction in a network of documents. %For the sake of simplicity, we only make use of the minimal attention unit of the \textit{Transformer}: the SDPA.

\section{Proposed Model}

We propose an algorithm for link prediction in text-attributed network. Our model is trained under a NE procedure, presented in Section \ref{sec:overall}. The optimization of the reconstruction error is performed via dot-product between contextual document representations $e^{v}_u$ and $e^{u}_v$. These embeddings are generated with a mutual attention mechanism over their textual contents only, described in Section \ref{sec:attention}. 

\subsection{Overall Optimization} \label{sec:overall}

The model takes as input a network of documents $G=(V,E,T)$, $T$ being the textual content of the documents. We precompute word embeddings $W$ and a normalized similarity measure between nodes $M$ designed from the adjacency matrix $A$ of the network. Each document $t_u$ is associated with a bag of word embeddings $W^{t_u}$ matrix. For any pair of node $(u,v) \in V^2$, mutual embeddings are generated with an asymmetric mutual attention function $f^A_{\Theta}$ for both documents given their bags of word embeddings $e^{v}_u = f^A_{\Theta}(W^{t_u}, W^{t_v})$ and $e^{u}_v = f^A_{\Theta}(W^{t_v}, W^{t_u})$. We define the unormalized similarity between the two nodes as the dot product of their mutual embeddings $e^{v}_u.e^{u}_v$. We aim at learning the parameters $\Theta$ by minimizing the KL divergence from the similarity distributions $M$ (from the graph) to that of the normalized distribution of the dot products between the mutual embeddings \cite{tsitsulin2018verse} (text associated to the nodes). We achieve this by employing noise-contrastive estimation \cite{tsitsulin2018verse}, minimizing the following objective function: $ J=-\sum_{(u,v) \in C} \Big( \log \sigma(e^{v}_u \cdot e^{u}_v) + \sum_{i=1}^k \mathbb{E}_{z\sim q} \big [ \log \sigma(-e^{z}_u \cdot e^{u}_z) \big ] \Big)$, where $\sigma$ is the sigmoid function. $C$ is a corpus of pairs of nodes generated by drawing uniformly existing links from the empirical distribution of links $M$. k negative nodes are uniformly drawn for each positive pair. To minimize this objective, we employ stochastic gradient descent using ADAM \cite{kingma2014adam}.

\subsection{Mutual Attention Mechanism} \label{sec:attention}

The role of $f^A_{\Theta}(W^{t_u}, W^{t_v})$ is to generate a contextual representation of $t_u$ given $t_v$. The parameters $\Theta$ we aim to learn are composed of three matrices $\Theta = \{P^Q, P^K, P^V\}$ of dimension $D \times D$ each. For all words of the target document $t_u$, we create queries $Q_u = W^{t_u} P^Q$. We similarly create keys and values from the contextual document $t_v$, such that $K_v = W^{t_v} P^Q$ and $V_v = W^{t_v} P^V$. Attention representations for each target word are then computed, following the SDPA formula: $ \text{SDPA}_{\Theta}(W^{t_u}, W^{t_v}) = \text{softmax}(\frac{Q_uK_v^T}{\sqrt{D}})V_v $. Note that $\text{SDPA}_{\Theta}(W^{t_u}, W^{t_v})$ has dimension $L \times D$, that is, we have a mutual attention representation of each word of document $t_u$ given $t_v$. Finally, the representation for document $t_u$ is obtained by averaging its word mutual attention vectors: $ e^{v}_u = f^A_{\Theta}(W^{t_u}, W^{t_v}) = \sum_{i=0}^L \text{SDPA}_{\Theta}(W^{t_u}, W^{t_v})_i $. Similarly, $e^{u}_v$ is generated by flipping indices $u$ and $v$. The intuition behind this model is that the matrices $P^Q$ and $P^K$ learn to project pairs of words that explain links in the network such that their dot-products produce large weights. $P^V$ is then meant to project the word vectors such that their average produces similar representations for nodes that are close in the network and dissimilar for nodes that are far in the network.    

\section{Experiments}

To assess the quality of our model, we perform two tasks of link prediction on a dataset of citation links between scientific abstracts: Cora \footnote{Get the data: https://linqs.soe.ucsc.edu/data}. The first prediction evaluation, called edges-hidden, consists in hiding a percentage of the links given a network of documents and measuring the ability of the model to predict higher scores to hidden links than to non-existing ones by computing the ROC AUC. The second evaluation, called nodes-hidden, consists in splitting the network into two unconnected networks, keeping a percentage of the nodes in the training network. %The model is then trained on the first graph and we measure its ability to predict links in the second graph using the ROC AUC. All scores are computed by dot-products of the representations produced by the algorithms.

We precompute on the full corpus word embeddings using \textit{GloVe} \cite{pennington2014glove} of dimension 256 with a co-occurrence threshold $x_\text{max}=10$, a window size $w=5$ and 50 epochs. We precompute \textit{LSI} \cite{deerwester1990ilsa} vectors of dimension 128. 
For the edge-hidden prediction task, we provide results performed by \textit{NetMF} with $k=10$ negative samples. \textit{TADW} is run with $20$ epochs and \modelname is performed with $k=1$ negative sample and $10^5$ sampled pairs of documents. The empirical similarity matrix between the nodes we chose is the normalized adjacency matrix. All produced representations are of dimension 256. %Note that \textit{NetMF} is unable to predict links for unseen documents. 

\subsection{Results}
\begin{table}[h]
\center
\caption{Edges-hidden link prediction ROC AUC}
\label{table:edges-hidden}
\begin{tabular}{l|ccccc}
\% of training data &10\%  &20\%  &30\%  &40\%  &50\%\\ \hline
%\textit{LSI} &- &- &- &- &79.5 \\
\textit{NeMF}  &59.0 &67.2 &77.5 &83.2 &87.2  \\ 
\textit{TADW}   &68.0 &82.0 &87.1 &\textbf{93.2} &\textbf{94.5} \\
\modelname &\textbf{82.3} &\textbf{87.1} &\textbf{88.6} &90.9 &91.0 \\
\end{tabular}
\end{table}

\begin{table}[h]
\center
\caption{Nodes-hidden link prediction ROC AUC}
\label{table:nodes-hidden}
\begin{tabular}{l|ccccc}
\% of training data &10\%  &20\%  &30\%  &40\%  &50\%\\ \hline
%\textit{LSI}  &- &- &- &- &79.5\\
\textit{TADW} &64.2 &\textbf{75.8} &\textbf{80.3} &\textbf{81.9} &\textbf{82.3} \\
\modelname &\textbf{69.4} &73.0 &75.4 &77.9 &78.6   \\
\end{tabular}
\end{table}

Tables \ref{table:edges-hidden} and \ref{table:nodes-hidden} show the results of our experiments. \modelname shows promising results for learning on a small percentage of training data on both evaluations. \textit{TADW} has better scores for nodes-hidden predictions which might be explained by the capacity of \textit{LSI} to learn discriminant features on a small dataset unlike \textit{GloVe}. In future work we would like to deal with bigger datasets from which word embedding methods might capture richer semantic information.

\bibliographystyle{ACM-Reference-Format}
\balance 
\bibliography{main}


\begin{thebibliography}{11}

%%% the Web address.

\ifx \showCODEN    \undefined \def \showCODEN     #1{\unskip}     \fi
\ifx \showDOI      \undefined \def \showDOI       #1{#1}\fi
\ifx \showISBNx    \undefined \def \showISBNx     #1{\unskip}     \fi
\ifx \showISBNxiii \undefined \def \showISBNxiii  #1{\unskip}     \fi
\ifx \showISSN     \undefined \def \showISSN      #1{\unskip}     \fi
\ifx \showLCCN     \undefined \def \showLCCN      #1{\unskip}     \fi
\ifx \shownote     \undefined \def \shownote      #1{#1}          \fi
\ifx \showarticletitle \undefined \def \showarticletitle #1{#1}   \fi
\ifx \showURL      \undefined \def \showURL       {\relax}        \fi
% invisible to TeX
\providecommand\bibfield[2]{#2}
\providecommand\bibinfo[2]{#2}
\providecommand\natexlab[1]{#1}
\providecommand\showeprint[2][]{arXiv:#2}

\bibitem[\protect\citeauthoryear{Deerwester, Dumais, Furnas, Landauer, and
  Harshman}{Deerwester et~al\mbox{.}}{1990}]%
        {deerwester1990ilsa}
\bibfield{author}{\bibinfo{person}{Scott Deerwester}, \bibinfo{person}{Susan~T.
  Dumais}, \bibinfo{person}{George~W. Furnas}, \bibinfo{person}{Thomas~K.
  Landauer}, {and} \bibinfo{person}{Richard Harshman}.}
  \bibinfo{year}{1990}\natexlab{}.
\newblock \showarticletitle{Indexing by latent semantic analysis}.
\newblock \bibinfo{journal}{\emph{JOURNAL OF THE AMERICAN SOCIETY FOR
  INFORMATION SCIENCE}} \bibinfo{volume}{41}, \bibinfo{number}{6}
  (\bibinfo{year}{1990}), \bibinfo{pages}{391--407}.
\newblock


\bibitem[\protect\citeauthoryear{Devlin, Chang, Lee, and Toutanova}{Devlin
  et~al\mbox{.}}{2018}]%
        {devlin2018bert}
\bibfield{author}{\bibinfo{person}{Jacob Devlin}, \bibinfo{person}{Ming-Wei
  Chang}, \bibinfo{person}{Kenton Lee}, {and} \bibinfo{person}{Kristina
  Toutanova}.} \bibinfo{year}{2018}\natexlab{}.
\newblock \showarticletitle{Bert: Pre-training of deep bidirectional
  transformers for language understanding}.
\newblock \bibinfo{journal}{\emph{arXiv preprint arXiv:1810.04805}}
  (\bibinfo{year}{2018}).
\newblock


\bibitem[\protect\citeauthoryear{Kingma and Ba}{Kingma and Ba}{2014}]%
        {kingma2014adam}
\bibfield{author}{\bibinfo{person}{Diederik~P Kingma} {and}
  \bibinfo{person}{Jimmy Ba}.} \bibinfo{year}{2014}\natexlab{}.
\newblock \showarticletitle{Adam: A method for stochastic optimization}.
\newblock \bibinfo{journal}{\emph{arXiv preprint arXiv:1412.6980}}
  (\bibinfo{year}{2014}).
\newblock


\bibitem[\protect\citeauthoryear{Mikolov, Chen, Corrado, and Dean}{Mikolov
  et~al\mbox{.}}{2013}]%
        {mikolov2013efficient}
\bibfield{author}{\bibinfo{person}{Tomas Mikolov}, \bibinfo{person}{Kai Chen},
  \bibinfo{person}{Greg Corrado}, {and} \bibinfo{person}{Jeffrey Dean}.}
  \bibinfo{year}{2013}\natexlab{}.
\newblock \showarticletitle{Efficient estimation of word representations in
  vector space}.
\newblock \bibinfo{journal}{\emph{arXiv preprint arXiv:1301.3781}}
  (\bibinfo{year}{2013}).
\newblock


\bibitem[\protect\citeauthoryear{Pennington, Socher, and Manning}{Pennington
  et~al\mbox{.}}{2014}]%
        {pennington2014glove}
\bibfield{author}{\bibinfo{person}{Jeffrey Pennington},
  \bibinfo{person}{Richard Socher}, {and} \bibinfo{person}{Christopher
  Manning}.} \bibinfo{year}{2014}\natexlab{}.
\newblock \showarticletitle{Glove: Global vectors for word representation}. In
  \bibinfo{booktitle}{\emph{Proceedings of the 2014 conference on empirical
  methods in natural language processing (EMNLP)}}.
  \bibinfo{pages}{1532--1543}.
\newblock


\bibitem[\protect\citeauthoryear{Perozzi, Al-Rfou, and Skiena}{Perozzi
  et~al\mbox{.}}{2014}]%
        {perozzi2014deepwalk}
\bibfield{author}{\bibinfo{person}{Bryan Perozzi}, \bibinfo{person}{Rami
  Al-Rfou}, {and} \bibinfo{person}{Steven Skiena}.}
  \bibinfo{year}{2014}\natexlab{}.
\newblock \showarticletitle{Deepwalk: Online learning of social
  representations}. In \bibinfo{booktitle}{\emph{Proceedings of the 20th ACM
  SIGKDD international conference on Knowledge discovery and data mining}}.
  ACM, \bibinfo{pages}{701--710}.
\newblock


\bibitem[\protect\citeauthoryear{Qiu, Dong, Ma, Li, Wang, and Tang}{Qiu
  et~al\mbox{.}}{2018}]%
        {qiu2018network}
\bibfield{author}{\bibinfo{person}{Jiezhong Qiu}, \bibinfo{person}{Yuxiao
  Dong}, \bibinfo{person}{Hao Ma}, \bibinfo{person}{Jian Li},
  \bibinfo{person}{Kuansan Wang}, {and} \bibinfo{person}{Jie Tang}.}
  \bibinfo{year}{2018}\natexlab{}.
\newblock \showarticletitle{Network embedding as matrix factorization: Unifying
  deepwalk, line, pte, and node2vec}. In \bibinfo{booktitle}{\emph{Proceedings
  of the Eleventh ACM International Conference on Web Search and Data Mining}}.
  ACM, \bibinfo{pages}{459--467}.
\newblock


\bibitem[\protect\citeauthoryear{Radford, Narasimhan, Salimans, and
  Sutskever}{Radford et~al\mbox{.}}{2018}]%
        {radford2018improving}
\bibfield{author}{\bibinfo{person}{Alec Radford}, \bibinfo{person}{Karthik
  Narasimhan}, \bibinfo{person}{Tim Salimans}, {and} \bibinfo{person}{Ilya
  Sutskever}.} \bibinfo{year}{2018}\natexlab{}.
\newblock \showarticletitle{Improving language understanding by generative
  pre-training}.
\newblock \bibinfo{journal}{\emph{URL https://s3-us-west-2. amazonaws.
  com/openai-assets/research-covers/language-unsupervised/language\_
  understanding\_paper. pdf}} (\bibinfo{year}{2018}).
\newblock


\bibitem[\protect\citeauthoryear{Tsitsulin, Mottin, Karras, and
  M{\"u}ller}{Tsitsulin et~al\mbox{.}}{2018}]%
        {tsitsulin2018verse}
\bibfield{author}{\bibinfo{person}{Anton Tsitsulin}, \bibinfo{person}{Davide
  Mottin}, \bibinfo{person}{Panagiotis Karras}, {and} \bibinfo{person}{Emmanuel
  M{\"u}ller}.} \bibinfo{year}{2018}\natexlab{}.
\newblock \showarticletitle{VERSE: Versatile Graph Embeddings from Similarity
  Measures}. In \bibinfo{booktitle}{\emph{Proceedings of the 2018 World Wide
  Web Conference on World Wide Web}}. International World Wide Web Conferences
  Steering Committee, \bibinfo{pages}{539--548}.
\newblock


\bibitem[\protect\citeauthoryear{Vaswani, Shazeer, Parmar, Uszkoreit, Jones,
  Gomez, Kaiser, and Polosukhin}{Vaswani et~al\mbox{.}}{2017}]%
        {vaswani2017attention}
\bibfield{author}{\bibinfo{person}{Ashish Vaswani}, \bibinfo{person}{Noam
  Shazeer}, \bibinfo{person}{Niki Parmar}, \bibinfo{person}{Jakob Uszkoreit},
  \bibinfo{person}{Llion Jones}, \bibinfo{person}{Aidan~N Gomez},
  \bibinfo{person}{{\L}ukasz Kaiser}, {and} \bibinfo{person}{Illia
  Polosukhin}.} \bibinfo{year}{2017}\natexlab{}.
\newblock \showarticletitle{Attention is all you need}. In
  \bibinfo{booktitle}{\emph{Advances in Neural Information Processing
  Systems}}. \bibinfo{pages}{5998--6008}.
\newblock


\bibitem[\protect\citeauthoryear{Yang, Liu, Zhao, Sun, and Chang}{Yang
  et~al\mbox{.}}{2015}]%
        {yang2015network}
\bibfield{author}{\bibinfo{person}{Cheng Yang}, \bibinfo{person}{Zhiyuan Liu},
  \bibinfo{person}{Deli Zhao}, \bibinfo{person}{Maosong Sun}, {and}
  \bibinfo{person}{Edward~Y Chang}.} \bibinfo{year}{2015}\natexlab{}.
\newblock \showarticletitle{Network representation learning with rich text
  information.}. In \bibinfo{booktitle}{\emph{IJCAI}}.
  \bibinfo{pages}{2111--2117}.
\newblock


\end{thebibliography}

\end{document}